\documentclass[letterpaper, 10 pt, conference]{ieeeconf}  % Comment this line out if you need a4paper

\IEEEoverridecommandlockouts                              % This command is only needed if 
\let\labelindent\relax % avoid IEEEtran error https://tex.stackexchange.com/questions/170772

\usepackage[bookmarks=true]{hyperref}
\usepackage{nrmac}
\usepackage{optidef}
\usepackage{svg}
\usepackage{booktabs}
\usepackage{multirow}
\usepackage{multicol}
\usepackage{mathptmx}
\usepackage{nicefrac}
\usepackage{subcaption}
\usepackage{mdframed}

\title{\LARGE \bf
An Approach for Generating Families of Energetically Optimal Gaits from Passive Dynamic Walking Gaits}

\author{Nelson Rosa$^{1}$ \and Bassel Katamish$^{2}$ \and Maximilian Raff$^{1}$ \and C. David Remy$^{1}$%
\thanks{*This work was funded by the Deutsche Forschungsgemeinschaft (DFG, German Research Foundation) – 501862165. It was further supported through the International Max Planck Research School for Intelligent Systems (IMPRS-IS) for Maximilian Raff and an Alexander von Humboldt fellowship to Nelson Rosa.}
\thanks{$^{1}$Nelson Rosa, Maximilian Raff, and C. David Remy are with the Institute for Nonlinear Mechanics, University of Stuttgart, Germany.
        {\tt\small \{nr, raff, remy\}@inm.uni-stuttgart.de}}%
\thanks{$^{2}$Bassel Katamish is with the Control Systems Group, Technical University of Berlin, Germany. {\tt \small katamish@scioi.de}}        
}

\usepackage{nccmath}
\usepackage{lmodern}     % set math font to Latin modern math
\usepackage[T1]{fontenc}
\DeclareSymbolFont{myletters}{OML}{ztmcm}{m}{it}
\DeclareMathSymbol{\uplambda}{\mathord}{myletters}{"15}

\newcommand{\Tr}{^{\mathop{\mathrm{T}}}}

\newcommand\inmfmtchar[1]{%
\ifx (#1(\else
\ifx )#1)\else
\ifx +#1+\else
\boldsymbol{#1}\fi\fi\fi
}

\usepackage{tokcycle, xcolor}
\Characterdirective{\addcytoks{\inmfmtchar{#1}}}
\Groupdirective{\processtoks{#1}}
\Macrodirective{\addcytoks{#1}}
\Spacedirective{\addcytoks{#1}}

\newcommand\inmfmt[1]{\tokcyclexpress{#1} \the\cytoks}

\newcommand{\deffunB}[1]{\expandafter\newcommand\csname #1\endcsname[1]{#1(##1)}}

\newcommand{\Q}{\mathcal{Q}}
\newcommand{\X}{\mathcal{X}}
\newcommand{\Sc}{\mathcal{S}}
\newcommand{\Sopt}{\Sc \times \Rl}

\newcommand{\B}{\mathfrak{B}}

\newcommand{\oc}{\tfrac{\partial J}{\partial c}\left(c\right)\Tr  + \begin{bmatrix} \pd{P}{c}\left(c\right)\Tr,  & \pd{\Ip}{c}\left(c\right)\Tr  \end{bmatrix} \lambda}

\newcommand{\cargs}{x_0, \tau, a}

\newcommandx{\optargs}[1][1=]{c#1, \lambda#1, p#1}

\newcommand{\dimx}{2n}
\newcommand{\dimp}{o}
\newcommand{\dima}{k}
\newcommand{\dimtau}{1}

\newcommand{\Rp}{\R^\dimp}
\newcommand{\Ra}{\R^\dima}
\newcommand{\Rtau}{\R}
\newcommand{\Rl}{\R^{\dimx + \dimp}}

\newcommand{\Rx}{\R^{2n}}
\newcommand{\Ru}{\R^{m}}

\newcommandx{\OP}[1][1=p]{\operatorname{OP}({#1})}
\newcommand{\passive}[1]{{#1}_\mathrm{pw}}
\newcommandx{\optimal}[2][2=p]{{#1}_{\OP[#2]}}
\newcommand{\desired}[1]{{#1}_\mathrm{des}}

\newcommandx{\Popt}[1][1=p]{\optimal{P}[#1]}

\newcommand{\slope}{\gamma}
\newcommand{\velocity}{v}

\newcommand{\vavg}{\velocity_\mathrm{avg}}

\newcommand{\Ip}{\Phi_p}

\newcommand{\fun}[3]{#1: #2 \to #3}
\newcommand{\pre}[1]{#1^{-1}(0)}

\newcommand{\Me}{M_\epsilon}

\newcommandx{\flow}[3][1=\tau, 2=, 3=(x_0)]{#2{\varphi}^{#1}_{a}#3}

\DeclareMathOperator{\flip}{flip}

\newmdenv[ %
frametitlerule=true, %
frametitlebackgroundcolor=gray!20, %
backgroundcolor=gray!10 %
]{infobox}

\newlength{\defparindent}
\svgpath{{Figures/}}
\graphicspath{{Figures/}}

\begin{document}

\maketitle
\thispagestyle{empty}
\pagestyle{empty}

%%%%%%%%%%%%%%%%%%%%%%%%%%%%%%%%%%%%%%%%%%%%%%%%%%%%%%%%%%%%%%%%%%%%%%%%%%%%%%%%
\begin{abstract}
For a class of biped robots with impulsive dynamics and a non-empty set of passive gaits (unactuated, periodic motions of the biped model), we present a method for computing continuous families of locally optimal gaits with respect to a class of commonly used energetic cost functions (e.g., the integral of torque-squared).  We compute these families using only the passive gaits of the biped, which are globally optimal gaits with respect to these cost functions.  Our approach fills in an important gap in the literature when computing a library of locally optimal gaits, which often do not make use of these globally optimal solutions as seed values.  We demonstrate our approach on a well-studied two-link biped model.
\end{abstract}

\section{Introduction}
\label{sec:intro}
The field of bipedal locomotion has made substantial gains in the past few decades with an eye (and leg) towards future commercialization.  At the core of these gains are gait generation algorithms that provide higher-level planners and controllers with reference trajectories for moving the robot from one location to the next.  While the goal is often to generate gaits that are energetically optimal in order to prolong onboard battery life, most approaches in the literature only provide local, typically point-wise, information of a biped's set of optimal gaits.  To the best of our knowledge, there are very few works that attempt to understand global properties of families of energetically optimal gaits.
\begin{figure}[th!]
    \centering
    \begin{subfigure}{0.35\columnwidth}
        \centering
        \includesvg[width=\columnwidth]{model}
        \caption{}
    \end{subfigure} \hfill
    \begin{subfigure}{0.6\columnwidth}
        \centering
        \includesvg[width=\columnwidth]{seed}
        \caption{}
        \vfill
        \includesvg[width=\columnwidth]{level-ground}
        \caption{}
    \end{subfigure} \\    
    \begin{subfigure}{\columnwidth}
        \centering
        \includegraphics[width=\columnwidth]{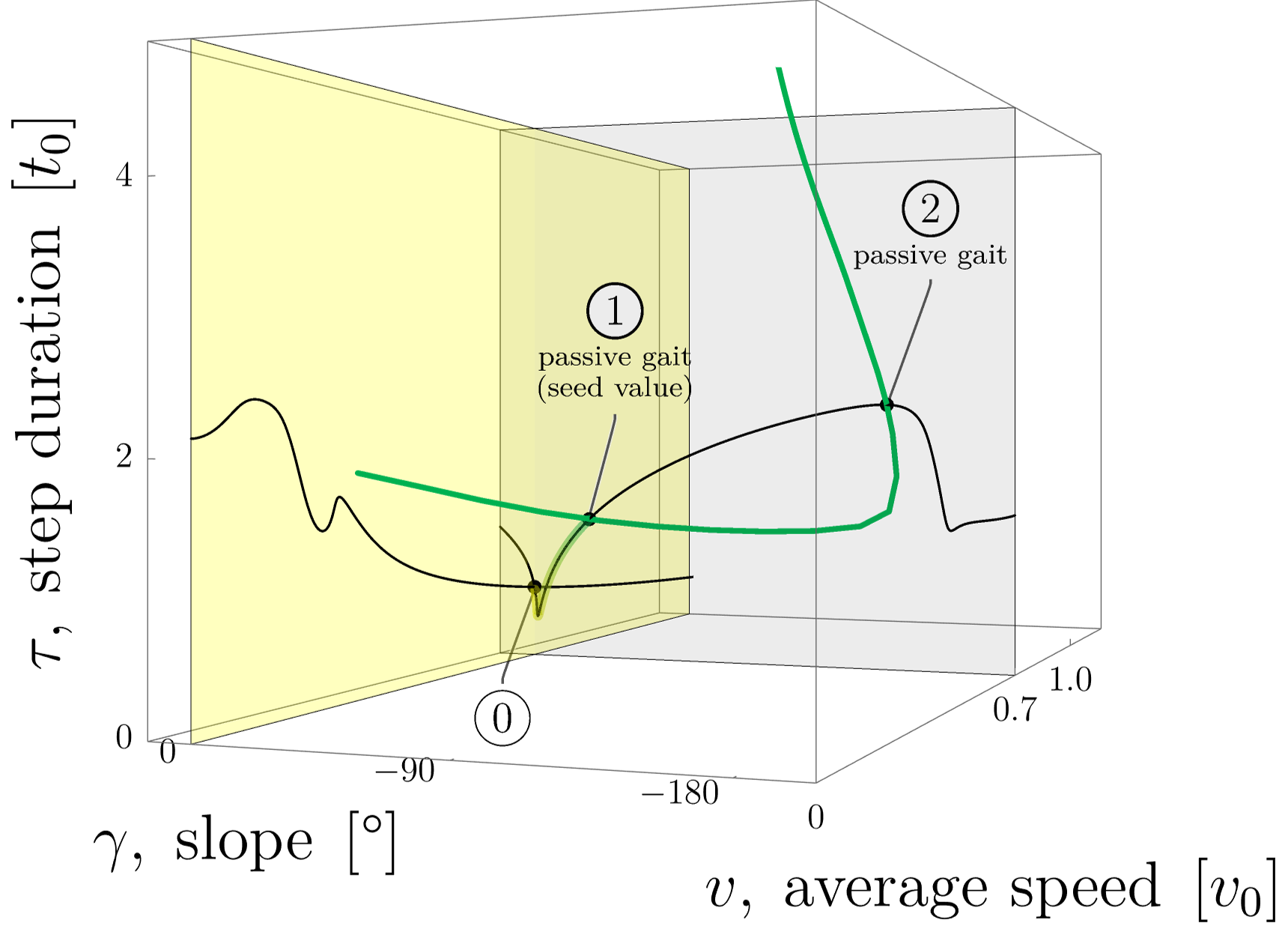}
        \caption{}
    \end{subfigure} \\    
    \begin{subfigure}{\columnwidth}
        \centering
        \includesvg[width=\columnwidth]{u-seed-to-level}
        \caption{}
    \end{subfigure}
    \caption{A demonstration of our approach on (a) a two-link biped robot with an actuator at the hip using (b) a passive (i.e., unactuated) dynamic walking motion of the biped model to compute continuous sets of energetically optimal actuated gaits, including (c) gaits that walk on level ground.  (d) Using a single passive gait as a seed value from a family of unactuated gaits (green curve), we are able to generate a curve of gaits with the same walking speed across a range of slopes (black curve in gray plane) and then switch to generating a curve of gaits that walk on level-ground across a range of walking speeds (black curve in yellow plane).  (e) Examples of locally optimal actuation profiles generated for a fixed speed and range of slopes in between gaits depicted in (b) and (c).  The example gaits are highlighted in (d) using a green to yellow color gradient between the seed value and level-ground walking gait.  The model is scaled with time measured in units of $t_0$, torque in units of $u_0$, and speed in units of $v_0$ (see Table~\ref{tab:parameters}).}
    \label{fig:slices3D}
\end{figure}

In this paper, we explore an outstanding problem in bipedal locomotion: what is the relationship between passive dynamic walking (PW, gaits that are capable of walking with zero actuation under the influence of gravity) and the set of actuated, energetically optimal, walking gaits in a biped's trajectory space (the set of all motions satisfying the biped's hybrid equations of motion)?  Our approach is unique in that our results capture the connectivity properties of the set of energetically optimal gaits and demonstrates an explicit connection between PW and actuated gaits that are locally optimal in a biped's trajectory space (see Figure~\ref{fig:slices3D}).

In particular, we demonstrate how to transform a parametric, equality-constrained optimization problem into an implicit function that defines a manifold of energetically optimal gaits (stationairy points of the parametric optimization problem) for bipeds modeled as impulsive dynamical systems with instantaneous impacts and parameterized open-loop control inputs.  The use of an implicit function to define a manifold of optimal gaits motivates our use of numerical continuation methods to trace a subset of these gaits across 1D slices on the manifold.  These slices are user-defined and can be used to generate curves of gaits across a range of operating points such as slopes and walking speeds (see Figure~\ref{fig:slices3D}).

The goal of our work is to demonstrate how to systematically generate families of optimal gaits and to analyze the effect of seeding these families with PW gaits, which for many common metrics of energetic efficiency are globally optimal solutions.  Our contributions are
\begin{enumerate}
    \item the conversion of a parametric optimization problem into an implicit function whose solution set can be traced using numerical continuation methods,
    \item an approach for generating a continuous family of energetically optimal gaits with respect to state, input parameters, and step duration, and
    \item the use of passive dynamic walking gaits as seed values to generate optimally actuated walking gaits.
\end{enumerate}

%%%%%%%%%%%%%%%%%%%%%%%%%%%%%%%%%%%%%%%%

%\subsection{Previous Work}
A common task in bipedal gait generation is to build a library of optimal gaits over a range of operating points, such as slopes and average walking velocities in a biped's trajectory space with respect to a parameterized input forcing function \cite{Gong2019, Xie2020, Reher2021, Raff2022}.

This work constructs a library of gaits that builds upon the methods and ideas in \cite{Raff2022, Rosa2021}, which treat the gait generation problem as tracing implicitly defined curves of periodic motions.  These curves can be traced using numerical continuation methods \cite{Allgower1990}, which are a class of algorithms for computing solutions of a system of parameterized equations.

A unique contribution of this work when compared to \cite{Raff2022} is the ability to directly trace optimal gaits for dissipative systems without having to first transform the model into an energetically conservative system.  Additionally, the results of \cite{Rosa2021} do not explicitly deal with how to generate optimal gaits.  This paper defines such a map for use in the algorithms outlined in \cite{Rosa2021}.

The standard approach for finding optimal gaits is to use a gradient-based optimization solver (e.g., Matlab's fmincon, SNOPT, or IPOPT) to find stationary points of the corresponding optimization problem.  Many works in this area focus on the problem formulation in an effort to encode a specific task \cite{Bessonnet2005, Ramezani2013} or demonstrate fast and reliable convergence to an optimal gait \cite{Hereid2018, Fevre2020}.  In certain cases, some frameworks have demonstrated that the generated gaits can be transferred to an experimental robot biped \mbox{(e.g., \cite{Ramezani2013, Hereid2018})}.  However, the energetic efficiency of these gaits still lags behind what many of the robot's biological counterparts can achieve \cite{Kim2017}.  This alone motivates the continued research in robotic bipedal gait generation.
% UPDATE: Kim2017 addresses citation feedback

While biped robots generally cannot walk as efficiently as, for example, a person, there exists machines that are capable of extremely efficient locomotion \cite{Bhounsule2014}.  The inspiration behind these designs stem from work in the passive dynamic walking literature \cite{McGeer1990a, Goswami1998, Garcia1998}, where simple models have been shown to walk down shallow inclines with no actuation.  More recently, methods for computing PW gaits for more complicated 2D and 3D biped models have been proposed~\cite{Rosa2021, Rosa2014a}.

This leads to an interesting gap in the research, where there is a disconnect between PW and the optimal gaits that many existing frameworks produce.  This motivates us to pose a variant of the optimal gait generation problem that captures aspects of the challenges presented in past works~\cite{Grizzle2014} and contributes results to include the use of passive gaits to generate a continuum of optimal gaits.

%%%%%%%%%%%%%%%%%%%%%%%%%%%%%%%%%%%%%%%%

%\subsection{Paper Outline}
In the remainder of this paper, we describe our biped modeling and cost function assumptions in Sections~\ref{sec:model}--\ref{sec:optimality}.  In Sections~\ref{sec:numerical}--\ref{sec:tracing}, we present the approach and apply it to a two-link biped model.  We end with a discussion of our results and conclude in Section~\ref{sec:conclusion}.

%%%%%%%%%%%%%%%%%%%%%%%%%%%%%%%%%%%%%%%%

\section{The Biped Model}
\label{sec:model}
We model a step of an $n$-degree-of-freedom biped with configuration $q \in \Q$, state $x = (q, \dot{q}) \in \X = T\Q \subset \Rx$, and step duration $\tau \in \Rtau$ as an impulsive dynamical system.  The resulting trajectories $x(t) \in \X$ for $0 \leq t \leq \tau$ capture the continuous motion of the biped pivoting about its stance foot and the collision of the swing foot with the ground as an instantaneous, plastic impact event at a discrete point in time.

There is no double support phase in our model.  During a foot-ground collision event, the stance foot prior to the event immediately breaks contact with the ground as the corresponding swing foot makes contact with the ground.  This leads to a discontinuous jump in the velocities of the biped at the time of impact \cite{Goswami1998, Ramezani2013, Hereid2018, Fevre2020, Westervelt2007}.

\subsection{The Impulsive Dynamics}
An impulsive dynamical system consists of continuous motion that satisfies $\dot{x}(t) = f(x(t), u(t)),$ where ${\dot{x}(t) = (\dot{q}(t), \ddot{q}(t)) \in T\X}$ is the time derivative of the state $x(t) \in \X$ at time $t$, $\fun{f}{\X \times \Ru}{T\X}$ is a vector field on $\X$, and $u(t) = [u_1(t), \ldots, u_i(t), \ldots, u_m(t)]^T \in \Ru$ is the control input at time $t$.  We define an individual input $u_i(t) \in \R$ ($1 \leq i \leq m$) in terms of $k_i \in \mathbb{N}$ basis functions and coefficients such that $u_i(t) = \sum_{j = 1}^{k_i} a_{i,j} \B_{i,j}(t),$ where, for a fixed $i$, $\B_{i,j}(t) \in \R$ and $a_{i,j} \in \R$ are the $k_i$ basis functions and coefficients, respectively, of $u_i(t)$ applied during the biped's continuous motion.

An example input for a biped with a single actuator (neglecting the subscript $i$) is a (half range) Fourier series with frequencies $w_j=2\pi j/\tau$ and $k$ coefficients
\begin{equation*}
u(t) = \frac{a_{1}}{2} + \sum_{j = 1}^{(k-1)/2} a_{2j} \cos\left(\omega_j t\right) + a_{2j+1} \sin\left(\omega_j t\right).
\label{eq:fourier}
\end{equation*}
The use of Fourier series to generate reference inputs has been used in the literature \cite{Iida2009}.  We implement the controller with $k = 3$ for our example biped.

The biped's continuous motion ends at a collision event.  A collision occurs whenever the time-state pair $(t, x(t))$ is an element of a guard set $S \subset \R \times \X$.  Given that collisions in our model occur at an instantaneous point in time, a jump map $\fun{\Delta}{\X}{\X}$ takes a pre-impact state $x(t^-)$ prior to a collision at time $t$ to a post-impact state $x(t^+)$ immediately after the collision such that $x(t^+) = \Delta(x(t^-)),$ where $t^-$ and $t^+$ represent the left- and right-sided limits of $x(t)$, respectively.  

\begin{mydef}
\label{def:S}
For $0 \leq t \leq \tau$, the \emph{impulsive dynamics} of a biped robot is the tuple $\Sigma = (\X, f, \Delta, S)$
\begin{equation}
\label{eq:S}
\Sigma : 
\begin{cases} 
    \begin{aligned}
    \dot{x}(t) &= f(x(t), u(t)) & (t, x(t^-)) \notin S, \\
    x(t^+) &= \Delta(x(t^-)) & (t, x(t^-)) \in S,
    \end{aligned}
\end{cases}
\end{equation}
where $f$, $\Delta$, $S$, etc., are as defined earlier in this section.
\end{mydef}

When the biped is modeled as a mechanical system, we can derive $f$ and $\Delta$ from the Euler-Lagrange equations (e.g., Appendix A of \cite{Rosa2021}).

\begin{myrem}
The definition of the guard set $S$ as a subset of an extended time-state space $\R \times \X$ unifies the autonomous \cite{Grizzle2014, Westervelt2007} and nonautonomous \cite{Rosa2021} impulsive dynamical system modeling found in the bipedal walking literature.
\end{myrem}

For trajectories satisfying $\Sigma$, the resulting step-to-step map $\flow \in \X$ gives the state of the robot at the end of a step
\begin{equation*}
\flow = x(\tau) = \Delta\left(x_0 + \ad{0}{\tau}{f(x(t), u(t))}{t}\right),
\label{eq:s2s}
\end{equation*}
where we parameterize trajectories with a post-impact state $x_0 \in \X$, a step duration $\tau$, and a control parameter vector $a \in \Ra$ ($\dima = \sum_{i = 1}^{m}k_{i}$) of coefficients of $u(t)$.

\begin{mydef}
\label{def:trajspace}
A point $c = (\cargs)$ in a biped's \emph{trajectory space} $\Sc = \X \times \Rtau \times \Ra$ defines the evolution of a step of a biped starting from post-impact state $x_0$ at time $t = 0$ and ending at post-impact state $\flow$ at $t = \tau$ under inputs $u(t)$ defined by a control parameter vector $a$.  If $c$ represents an unactuated trajectory, then we must have $a = 0$ for $u(t) = 0$ during a step.
\end{mydef}

\subsection{Modeling Assumptions}
\label{ssec:step}
In order to generate a continuous set of optimal gaits, we require trajectories to be twice differentiable with respect to points $(\cargs)$ in the trajectory space $\Sc$ of Definition~\ref{def:trajspace}.

\begin{myas}
\label{as:trajectories}
We assume
\begin{enumerate}[label=A\theenumi]
\item \label{as:trans}
for a given guard set $S = \{(t, x(t)): \varphi(t, x(t)) = 0\}$, where $\fun{\varphi}{\R \times \X}{\R}$ defines the switching surface, a trajectory $x(t)$ intersects the surface $\pre{\varphi}$ transversally at a foot-ground collision event,

\item \label{as:diff}
the functions ${u(t) \in \Ru}$, ${f(x, u) \in T\X}$, and ${\Delta(x) \in \X}$ are twice continuously differentiable with respect to time~$t \in \R$, state $x \in \X$, and controls parameters $a \in \Ra$ at all points in $\Sc$, and

\item \label{as:bs}
the biped has bilateral symmetry (i.e., the robot's ``left'' and ``right'' sides are mirror images of each other).
\end{enumerate}
\end{myas}
Assumption~\ref{as:trans} ensures that trajectories do not trigger multiple foot-ground collisions in an infinitesimally small amount of time.  Assumption~\ref{as:diff} ensures that solutions to $\Sigma$ of Equation~\ref{eq:S} exist, are unique, have finite left- and right-sided limits for all points on the interval of time considered, are discontinuous only at switching time $t = \tau$, 
and have a continuous dependence on and are differentiable with respect to the parameters of the system, $x_0$, $\tau$, and $a$ \cite{Simeonov1988}.  Finally, Assumptions~\ref{as:bs} focuses our study to half-strides of bipeds with symmetric gaits.

\section{Optimality in the Trajectory Space}
\label{sec:optimality}
A common task in bipedal gait generation is to find a set of optimal periodic gaits over a range of operating points. 

\subsection{Periodic Trajectories and Operating Points of Interest}
\begin{mydef}
\label{def:P}
A point $c = (\cargs)$ in the trajectory space $\Sc$ is a \emph{gait} of the biped if it is a root of the \emph{periodicity map} $\fun{P}{\Sc}{\Rx}$ such that $P(c) = \flow - \flip(x_0) = 0$, where $\flip$ maps $x_0$ to its ``mirrored'' state (i.e., the biped at the same position, but with the legs flipped).
\end{mydef}

\begin{mydef}
\label{def:setP}
The set $\pre{P}$ is the \emph{set of all gaits}, where the notation $\pre{P}$ is the set $\{c \in \Sc: P(c) = 0\}$.  The set $\passive{P} = \{(\cargs) \in \pre{P}: a = 0 \}$ is the \emph{set of all passive gaits}.
\end{mydef}

\begin{mydef}
\label{def:Phi}
A gait $c \in \pre{P}$ satisfies a vector of user-defined \emph{operating points} $p \in \Rp$ ($\dimp \leq \dima + \dimtau$) if it is the root of a twice-differentiable map $\fun{\Ip}{\Sc}{\Rp}$ such that ${\Ip(c) = p_\mathrm{act}(c) - p}$, where $p_\mathrm{act}(c) \in \Rp$ is a vector of values derived from gait $c$ that we want equal to the operating point.
\end{mydef}

A common pair of operating points are the biped's average walking speed and the incline of the walking surface \cite{Gong2019, Xie2020, Reher2021, Raff2022}.  The resulting constraint would be $\Ip(c) = [\slope(c) - \desired{\slope}, \vavg(c) - \desired{\velocity}]\Tr = 0$, where ${\fun{\slope}{\Sc}{\R}}$ and ${\fun{\vavg}{\Sc}{\R}}$ are a gait's slope and average walking speed, respectively, and $\desired{\slope}$ and $\desired{\velocity}$ are the operating points such that $p = [\desired{\slope}, \desired{\velocity}]\Tr$.

\begin{myrem}
Some readers may be surprised to see slope $\gamma$ as a function of $c$, which implies that slope depends on the step duration $\tau$.
In related work on PW, $\gamma$ is typically considered to be an independent variable and $\tau$ is determined by the time of foot strike.
Here, however, we switch the roles of the dependent and independent variables, where collision happens after a predefined time $\tau$, not at a predefined state, and slope $\gamma(c)$ is a dependent variable obtained from the robot's configuration at time $\tau$.  While this is not a representation of physical over-ground walking, the two views result in the same motion, as soon as the constraint in $\Ip(c)$ is introduced.
\end{myrem}

\subsection{Cost Functions of Interest}
In the space of trajectories $\Sc$, we want gaits that are optimal with respect to a twice-differentiable cost function $\fun{J}{\Sc}{\R}$ across a continuous range of user-defined operating points $p$.  As there are several metrics for measuring the energetic efficiency of a gait, we let the cost function be user-defined under the following assumption.

\begin{myas}
\label{as:J}
For a given twice-differentiable cost function $\fun{J}{\Sc}{\R}$, we assume passive gaits are global minima  of $J$.  That is for all passive gaits $\passive{c} \in \passive{P}$, $J(\passive{c}) \leq J(c)$ for all $c \in \Sc$.
\end{myas}

The most common energetic cost functions in legged locomotion (e.g, integral of squared torque \cite{Westervelt2007, Bessonnet2005, Fevre2020} and positive work \cite{Ramezani2013}) trivially satisfy this criterion as these cost functions are non-negative and are zero whenever the torques are zero.

%%%%%%%%%%%%%%%%%%%%%%%%%%%%%%%%%%%%%%%%

\section{Numerical Continuation of Optimal Gaits}
\label{sec:numerical}
We now present our primary contributions, where given the tuple $(\Sigma, P, \passive{P}, \Ip, J)$ of Definitions~\ref{def:S}--\ref{def:Phi} and Assumptions~\ref{as:trajectories}--\ref{as:J} and a non-empty set $\passive{P}$ of passive gaits, we find solutions to the following optimization problem
\begin{mini*}
        {c}{J(c)}{\label{opti:pop}}{\OP:}
        \addConstraint{P(c)}{= 0}
        \addConstraint{\Ip(c)}{= 0}
\end{mini*}
using numerical continuation methods.

%%%%%%%%%%%%%%%%%%%%%%%%%%%%%%%%%%%%%%%%

\subsection{Optimal Trajectories as Roots of a Map}
Given the optimization problem $\OP$, a point $c \in \Sc$ is an \emph{optimal gait} if $c$ satisfies the first-order optimality conditions (FOC) \cite{Nocedal1999}
\begin{equation}
\label{eq:oc}
\begin{gathered}
    P(c) = 0, \quad \Ip(c) = 0 \\ \oc = 0,
\end{gathered}
\end{equation}
where $\lambda \in \Rl$ is a vector of Lagrange multipliers.

This leads to the set of optimal gaits of $\OP$ as also being the roots of the map $\fun{\Popt}{\Sopt}{\R^{4n + \dima + \dimp + \dimtau}}$, where
\begin{equation*}
\begin{split}
\Popt(c, \lambda) &= \begin{bmatrix} P(c) \\ \Ip(c) \\ \oc \end{bmatrix}.
\end{split}
\label{eq:Popt}
\end{equation*}

\begin{mydef}
\label{def:mani}
\newcommand{\copt}{c^\ast}
\newcommand{\Pset}{\pre{M}}
\newcommand{\Jac}{\pd{M}{c}(\copt)}
Let $\fun{M}{\R^b}{\R^a}$ ($a \leq b$) be a continuously differentiable map and the point $c^* \in \pre{M}$ a root of the map.  If $\Jac$ has maximal rank, i.e., ${\rank\left( \Jac \right) = a}$, then $\copt$ is a \emph{regular point} of $\Pset$ (and a \emph{singular point} otherwise).  Regular points of $M$ have the following properties \cite{Spivak1965}:
\begin{enumerate}
    \item the tangent space of $\Pset$ at $\copt$, $T_{\copt} \Pset$, is equal to the null space of $\Jac$, and
    \item there exists a neighborhood of regular points containing $\copt$ that form a differentiable manifold in $\Pset$ of dimension $b - a$.
\end{enumerate}
\end{mydef}
Stated differently, if $c^\ast$ is a regular point of $M$, then we are guaranteed the existence of a $(b - a)$-dimensional solution family (specifically, a differentiable manifold) of roots of $M$ in a neighborhood of $c^\ast$.  For the regular points of $\Popt$, we have the following result.

\begin{myprop}
\label{prop:iso}
If the pair $(c^\ast, \lambda^\ast)$ is a regular point of $\Popt$ and a solution of $\OP$ for fixed $p$, then it is an isolated point in $\pre{\Popt}$.
\end{myprop}
\begin{proof}
The tangent space $T_{(c^\ast, \lambda^\ast)}\pre{\Popt}$ represents all of the feasible directions we can move in without violating the FOC (Equation~\ref{eq:oc}).  At a regular point, the tangent space equals the null space.  The null space is empty because the Jacobian of $\Popt$ is square and has maximal rank.  Hence, $(c^\ast, \lambda^\ast)$ is an isolated point in $\pre{\Popt}$.
\end{proof}
In the context of optimization, the gait $c^\ast$ is a strict local extremizer of $\OP$ at a fixed value of $p$.  This is a desirable property when numerically searching for a single locally optimal gait.  However, this is not a desirable feature for numerical continuation as the regular points for a map with fixed $p$ do not form a continuous family of gaits.

%%%%%%%%%%%%%%%%%%%%%%%%%%%%%%%%%%%%%%%%

\subsection{Passive Gaits as Seed Values}
We now generate a curve of optimal points in $\Sopt$ with the properties that the curve 1) starts from a passive gait $\passive{c} \in \passive{P}$, and, if a path exists, 2) intersects a gait $c$ such that $p_\text{act}(c) = p_\text{des}$, where $p_\text{des}$ is a desired operating point and $p_\text{act}(c)$ are the actual values for the gait. 
We accomplish this by letting the operating point $p$ be a function of a 1D parameter $\epsilon \in \R$ such that ${p := p(\epsilon) = (1-\epsilon) \desired{p} + \epsilon p_\text{act}(\passive{c})}$.  After plugging $p(\epsilon)$ into $\Popt$ and simplifying, we arrive at a global-homotopy-inspired map $\Me$ \cite{Allgower1990}
\begin{align}
\label{eq:Me}
    \Me(c, \lambda) &= \Popt[\desired{p}](c, \lambda) - \epsilon \Popt[\desired{p}](\passive{c}, \passive{\lambda}),
\end{align}
where $\epsilon \in \R$ is the homotopy parameter, and $\lambda \in \Rl$ and $\passive{\lambda} \in \Rl$ are Lagrange multipliers.  The regular points of the map implicitly define a curve with the properties mentioned earlier.  For $\epsilon = 1$, a passive gait is a root of $\Me$ and serves as our seed value.  At $\epsilon = 0$, a root of $\Me$ must be a gait that satisfies the operating point $\desired{p}$.  Furthermore, every point $(c_\epsilon,\lambda_\epsilon) \in \pre{\Me}$ on the curve is a periodic motion that is optimal with respect to its operating point~$p(\epsilon)$.

Given the map $\Me$ of Equation~\eqref{eq:Me}, we can apply the algorithms in \cite{Rosa2021} to trace curves of optimal gaits.  We provide a Mathematica implementation \cite{Rosa2023a}, which implements the pseudo-arclength continuation of \cite{Rosa2021} with some modifications.  Specifically, tangent vectors to the curve are computed internally because all points of the map are regular points, and we have incorporated an adaptive step-size scheme when tracing a curve in $\pre{\Me}$ (see \cite{Allgower1990} for details).

%%%%%%%%%%%%%%%%%%%%%%%%%%%%%%%%%%%%%%%%

\section{Tracing Continuous Families of Optimal Gaits of the Compass Gait Walker}
\label{sec:tracing}
We demonstrate our approach using the compass-gait walker \cite{Goswami1998}.  The biped (Figure~\ref{fig:slices3D}) consists of two legs of length $\ell_0$ and point masses $m_\ell$ on the legs and $m_H$ at the hip.  The external forces are gravity $g = [0, \; -g_0]\Tr \in \R^2$ and an actuator $u(t) \in \Ru$ at the hip ($m = 1$).  The slope $\slope$ is defined relative to level ground and for positive velocities divides the biped's gaits into four types: 1) uphill brachiation ($\slope \leq -180^\circ$), 2) downhill brachiation ($-180^\circ < \slope \leq -90^\circ$), 3) downhill walking ($-90^\circ < \slope < 0^\circ$), and 4) uphill walking (${0^\circ \leq \slope < 90^\circ}$).  We show example motions in the next section.

We can describe the motion using a set of minimal coordinates, where the initial state of the robot ${x_0 = [q_1, q_2, \dot{q}_1, \dot{q}_2]\Tr \in \X \subset \Rx}$ for an $n = 2$-degree-of-freedom system given that one foot is always in contact with the ground (Section~\ref{sec:model}).  It is straightforward to map the minimal coordinates to the coordinates depicted in Figure~\ref{fig:slices3D}. 
 Additionally, the model is nondimensionalized with physical and scaling parameters listed in Table~\ref{tab:parameters}.
\begin{table}[t]
    \centering
    \renewcommand*{\arraystretch}{1.2}
    \begin{tabular}{ccccc}
    \toprule
    \multicolumn{3}{c}{physical parameters} & \multicolumn{2}{c}{scaling parameters} \\
    \cmidrule(lr){1-3}\cmidrule(lr){4-5}
    quantity & value & units & quantity & value \\
    \cmidrule(lr){1-3}\cmidrule(lr){4-5}
    $\ell_0$ & 1 & $\unit{m}$ & $m_0$ & $2 m_\ell + m_H$  \\
    $b$ & 5 & $\unit{m}$ & $t_0$ & $\sqrt{\nicefrac{\ell_0}{g_0}}$ \\
    $m_H$ & 10 & $\unit{kg}$ & $v_0$ & $\nicefrac{\ell_0}{t_0}$ \\
    $m_\ell$ & 1 & $\unit{kg}$ & $u_0$ & $\nicefrac{m_0 \ell_0^2}{t_0^2}$ \\
    $g_0$ & $9.81$ & $\unit{N}$ & $J_0$ & $u_0^2 t_0$ \\
    \bottomrule
    \end{tabular}
    \caption{The model parameters of the compass-gait walker.}
    \label{tab:parameters}
\end{table}

For the biped model, the task is to generate a library of gaits that can walk on various slopes $\slope$ and average walking speeds $\vavg$ along the slope.  The gaits must minimize the integral of the squared torques used during a step ${J = \ad{0}{\tau}{u(t)\Tr u(t)}{t}}$.  The library must also have gaits that walk on level ground with a desired operating point of ${\desired{\slope} = 0}$, so that we can further explore the relationship between energetically efficient level-ground walking and passive dynamic walking gaits.

In computing the library of gaits, we compute passive gaits for the biped using the methods described in \cite{Rosa2021}.  We then choose a seed value and compute $25$ points to the left and to the right of the seed value for a total of $51$ points on the curve.  The curve is computed using an ``adaptive step by asymptotic expansion'' as presented in \cite{Allgower1990} with an initial step size of $h_0 = \pm 0.02$.  The two pieces of a curve are computed in parallel across two processors.  For the figures in this section, we did a second run of the data using a finer mesh with a fixed step size.  We redid the data with a finer mesh in order to more accurately compute summary statistics of the cost along a curve.

Overall, we use our algorithm to compute two curves: a curve where gaits have the same velocity and a curve where the gaits have the same slope.  The coarse mesh for the constant-velocity curve took \qty{20}{\minute} to compute in parallel while the constant-slope curve took \qty{32}{\minute}.  We used these curves to guide us in determining the range of slopes and velocities for the finer meshes, which collectively took \qty{11}{\hour} to compute across four processor cores.

Finally, all trajectories $c = (\cargs)$ and their cost are computed in Mathematica using an explicit fourth-order Runge-Kutta scheme with a fixed step size of $\nicefrac{\tau}{30}$ on a Lenovo L380 Yoga
with an Intel i7-8550U CPU.  The use of a fixed integration step size is not necessary.  In principle, any ODE solver can be used to solve the equations of motion.  In Mathematica 11.1+, the built-in ODE solver can take hours to compute a single hybrid trajectory with an adaptive step size.  We switched to a fixed step size as a workaround.  A trajectory is divided into 30 time steps as we found it to be a good trade-off between speed and accuracy when compared to the trajectory computed with an adaptive step size.

\subsection{An Example Constant-Velocity Slice of Optimal Gaits}
\label{ssec:velocity}
\begin{figure}
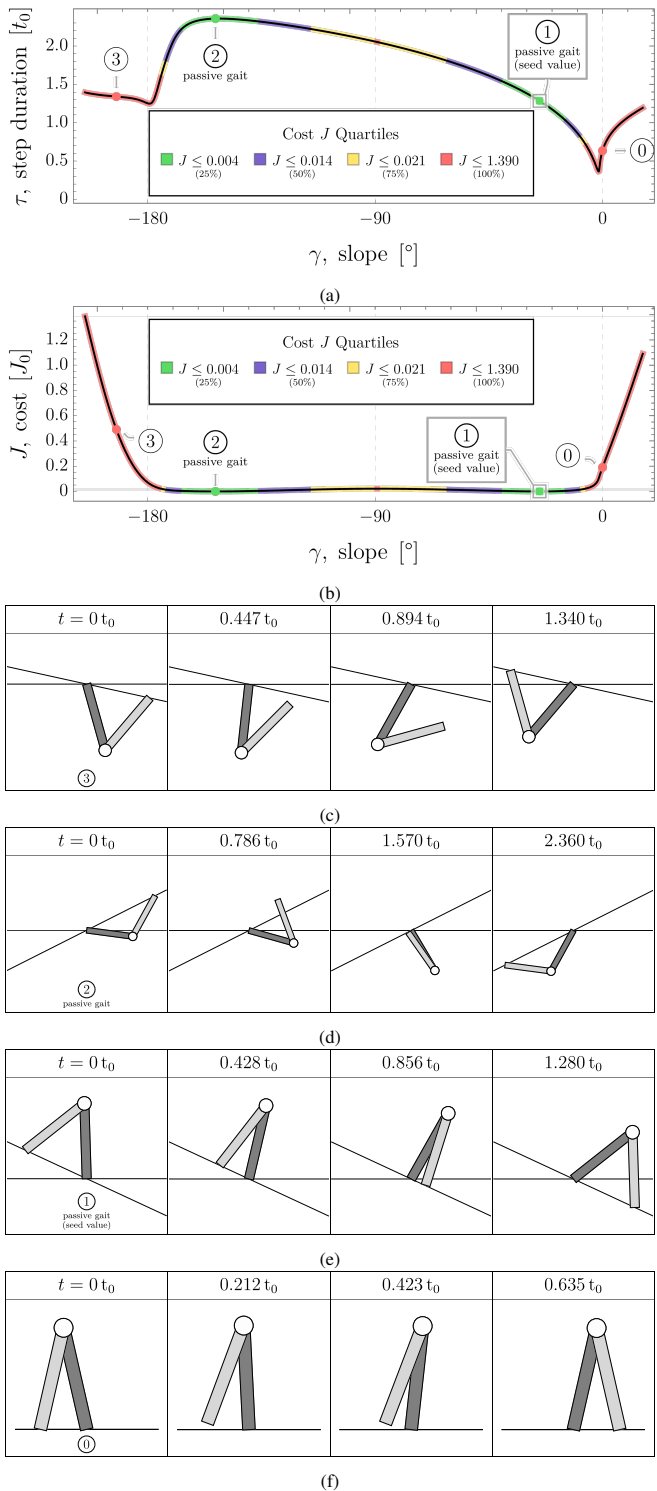

    \centering
    \begin{subfigure}{\columnwidth}
        \centering
        \includesvg[width=\columnwidth]{velocity-slice-time}
        \caption{}
    \end{subfigure} \\   
    \begin{subfigure}{\columnwidth}
        \centering
        \includesvg[width=\columnwidth]{velocity-slice-cost}
        \caption{}
    \end{subfigure} \\  
    \begin{subfigure}{\columnwidth}
        \centering
        \includesvg[width=\columnwidth]{cgw-opt-F-2-Z_brach_up.svg}
        \caption{}
    \end{subfigure} \\
    \begin{subfigure}{\columnwidth}
        \centering
        \includesvg[width=\columnwidth]{cgw-opt-F-2-Z_brach_down_pass.svg}
        \caption{}
    \end{subfigure} \\   
    \begin{subfigure}{\columnwidth}
        \centering
        \includesvg[width=\columnwidth]{cgw-opt-F-2-Z_walk_down_pass.svg}
        \caption{}
    \end{subfigure} \\ 
    \begin{subfigure}{\columnwidth}
        \centering
        \includesvg[width=\columnwidth]{cgw-opt-F-2-Z_walk_up.svg}
        \caption{}
    \end{subfigure}
  \caption{
  (a) A constant-velocity slice of optimal gaits projected onto a slope-step-duration subspace; the seed value is labeled with a (1). (b) The optimal cost as a function of slope along the curve in the constant-velocity slice; the first through fourth quartiles define the color coding of this plot (see inset legend). (c)--(f) Example gait motions of labeled points in the plots.}
  \label{fig:velocity}
\end{figure}
Figures~\ref{fig:velocity} show a constant-velocity slice of optimal gaits for the compass-gait walker projected onto slope-step-duration and slope-cost subspaces in Figures~\ref{fig:velocity}(a) and (b), respectively.  Every point on the curve represents an optimal gait with an average walking speed of $\vavg = \qty{0.7}{v_0}$.  The only form of actuation is at the hip, which outputs a Fourier series summed up to the first harmonic
\begin{equation}
u(t) = \frac{a_1}{2} + a_2 \cos\left(2\pi\frac{t}{\tau}\right) + a_3 \sin\left(2\pi\frac{t}{\tau}\right).
\label{eq:uhip}
\end{equation}
The number of control points is $k = 3$ with the control parameter vector $a = [a_1, a_2, a_3]\Tr$.

For this slice, the operating point $p$ as a function of $\epsilon$ is
\begin{equation}
    p := p(\epsilon) = 
    \begin{bmatrix} \slope \\ \vavg \end{bmatrix} = 
    \begin{bmatrix}
        (1 - \epsilon) \desired{\slope} + \epsilon \slope(\passive{c}) \\ \vavg(\passive{c}),
    \end{bmatrix}    
    \label{eq:p-velocity}
\end{equation}
where $\desired{\slope} = 0$ and the functions $\slope(c)$ and $\vavg(c)$ compute the gait's actual slope and velocity, respectively.

The seed value used to trace the curve is labeled (1), which is a passive gait of the biped.  The seed has a slope of $\qty{-24.9}{\degree}$ and a step duration of $\qty{1.28}{t_0}$.  We initially computed 25 points to the left and right of the curve for a total of 51 gaits.  The computed gaits from this initial data set have a range of slopes from $\qtyrange{-190}{15}{\degree}$ and step durations from $\qtyrange{0.37}{2.35}{t_0}$.

Figure~\ref{fig:velocity} plots the resulting curve using a finer mesh of 2209 points in total over a similar range of slopes.  In particular, Figure~\ref{fig:velocity}(a) plots the relationship between slope and step duration, and Figure~\ref{fig:velocity}(b) plots the cost of each gait along the curve.  The four colors partitioning the curve encode summary statistics of the first through fourth quartiles of the cost function along the curve [see inset of  Figure~\ref{fig:velocity}(b)].  For example, $\qty{25}{\percent}$ of the gaits computed (the first quartile) on the curve have a cost of at most $J = \qty{0.004}{J_0}$.  These gaits are highlighted in green.  The cost is relatively flat for downhill gaits and increases quickly as the gaits start to walk uphill.  The actuator in these uphill regions has to compensate for gravity as an opposing force to the direction of travel.  Example motions of points on the curve are depicted in Figure~\ref{fig:velocity}(c)--(f), which show brachiating and walking gaits that locomote up and downhill.

An interesting feature of this curve is that it intersects another globally optimal gait labeled (2) in the plots.  This highlights a strength of our approach in that optimizing with respect to step duration in addition to state and control parameters enables us to find the best possible gait for a given slope and velocity.  It further supports past trends of finding lower cost gaits when impact times are free parameters in bipedal gait optimization problems \cite{Ponton2018}.

%%%%%%%%%%%%%%%%%%%%%%%%%%%%%%%%%%%%%%%%

\subsection{Growing the Library of Gaits Along Different Slices}
\label{ssec:slope}
\begin{figure*}[t]
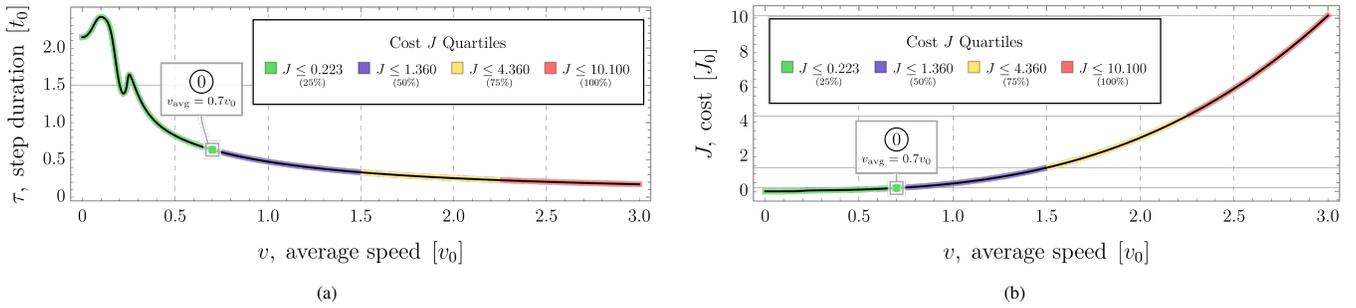

    \centering
    \begin{subfigure}{\columnwidth}
        \centering
        \includesvg[width=\columnwidth]{slope-slice-time}
        \caption{}
    \end{subfigure} \hfill
    \begin{subfigure}{\columnwidth}
        \centering
        \includesvg[width=\columnwidth]{slope-slice-cost}
        \caption{}
    \end{subfigure}
  \caption{
  (a) A constant-slope slice of optimal gaits projected onto a slope-step-duration subspace; the level-ground gait from the constant-velocity slice of gaits of Figure~\ref{fig:velocity} is labeled with a (0). (b) The optimal cost as a function of slope along the curve in the constant-velocity slice; the first through fourth quartiles define the color coding of this plot (see inset legend).}
  \label{fig:slope}
\end{figure*}
We can grow the library of optimal gaits along other slices of interest using our current set of passive and constant-velocity slice of gaits.  For example, the constant-velocity curve of gaits has an actuated gait $c^*$ that we can use to generate a curve of level-ground walking gaits with the input of Equation~\eqref{eq:uhip} and operating point
\begin{equation}
    p(\epsilon) = 
    \begin{bmatrix} \slope \\ \vavg \end{bmatrix} = 
    \begin{bmatrix}
        \slope(c^*) \\
        (1 - \epsilon) \desired{\velocity} + \epsilon \vavg(c^*)
    \end{bmatrix},
    \label{eq:p-slope}
\end{equation}
where we use the same notation as in Equation~\eqref{eq:p-velocity} and set $\desired{\velocity} = \qty{3}{v_0}$.

The gait $c^* = (\cargs)$ used to trace the curve is labeled (0) and is taken from the constant-velocity curve of Figure~\ref{fig:velocity}.  The gait is actuated ($a \neq 0$), has a slope of $\qty{0}{\degree}$, velocity of $\qty{0.7}{v_0}$, and a step duration of $\qty{0.64}{t_0}$.  As in the previous section, we computed a total of 51 points on the curve with an adaptive step-size.  The computed gaits from this data set have a range of velocities from $\qtyrange{0.16}{2.1}{v_0}$ and step durations from $\qtyrange{0.24}{2.2}{t_0}$.

Figure~\ref{fig:slope} plots the resulting curve using a finer mesh of 3001 points over a similar range of speeds and step durations. Figure~\ref{fig:velocity}(a) plots the relationship between walking speed and step duration and Figure~\ref{fig:slope}(b) plots the cost of each gait along the curve.

%%%%%%%%%%%%%%%%%%%%%%%%%%%%%%%%%%%%%%%%

\section{Discussion and Conclusion}
\label{sec:conclusion}
%% What we have done
The goal of our work was to connect passive motions of legged systems to families of energetically optimal actuated gaits.  The key idea behind our proposed approach was to construct a map whose roots are solutions to a family of optimization problems that are parameterized by a continuous range of operating points.  We traced the resulting zero set of this map using numerical continuation methods.  For the example of a two-link biped, we demonstrated the generation of two such curves of optimal gaits: one consisting of gaits with the same constant walking speed parameterized by slope, and the other consisting of gaits that all walk on level ground parameterized by speed.

%% Why it is important
The explicit use of passive gaits as starting points has the advantage that it results in a continuous set of optimal gaits that are derived from seed values that are optimally exploiting the natural mechanical dynamics and that are globally optimal for many energetic cost functions, including the most commonly used ones: positive mechanical work and integral of torque-squared.  The continuity property of our approach is an important contribution towards a more global understanding of where connected sets of optimal gaits exist in a legged systems' trajectory space

%% Potential for optimization
For future work, we want to further explore questions that are of interest in the optimization and bipedal robotics communities.  For example, in addition to Fourier series, B\'{e}zier and B-Splines curves are also a popular choice for actuation signals in the literature that are written in terms of basis functions.  Given that a passive gait does not change its trajectory or cost when embedded in these different parameter spaces, how does the cost landscape change for the connected sets of gaits with respect to the actuation strategy and the number of control points used?  

There is also a need to compare our approach to state-of-the-art methods and to better understand when continuation methods are a better choice over standard optimization-based methods.  This includes extending our approach to handle inequality constraints as in \cite{Rosa2021}.  We leave these avenues of study to future work.

%% Broad final paragraph and positive outlook
Finally, the ability to continuously connect passive motions that are solely based on the natural mechanical dynamics of a system to energetically optimal actuated gaits that can be observed on level ground has the potential to become a valuable tool in the study of the gait itself.  For example, we can use it to study qualitative trends as gaits change continuously, it can guide our understanding of what fundamental properties make an optimal gait optimal, and it can help us answer the question of why certain motion patterns emerge in the gaits of humans and animals.

%\addtolength{\textheight}{-9cm}   % This command serves to balance the column lengths
                                  % on the last page of the document manually. It shortens
                                  % the textheight of the last page by a suitable amount.
                                  % This command does not take effect until the next page
                                  % so it should come on the page before the last. Make
                                  % sure that you do not shorten the textheight too much.

%%%%%%%%%%%%%%%%%%%%%%%%%%%%%%%%%%%%%%%%%%%%%%%%%%%%%%%%%%%%%%%%%%%%%%%%%%%%%%%%

%%%%%%%%%%%%%%%%%%%%%%%%%%%%%%%%%%%%%%%%%%%%%%%%%%%%%%%%%%%%%%%%%%%%%%%%%%%%%%%%

%%%%%%%%%%%%%%%%%%%%%%%%%%%%%%%%%%%%%%%%%%%%%%%%%%%%%%%%%%%%%%%%%%%%%%%%%%%%%%%%
%\section*{Appendix}

%\section*{Acknowledgment}

\bibliographystyle{IEEEtran}
\bibliography{MyBib}
\end{document}